\documentclass[final]{svjour3}

\usepackage{graphics}
\usepackage{color}
\usepackage{graphicx}
\usepackage{tabularx}
\usepackage{url}
\usepackage{multicol}
\usepackage{amssymb,amsfonts,textcomp}
\usepackage{algorithm}
\usepackage{algorithmicx}
\usepackage{algpseudocode}
\usepackage[authoryear,semicolon,sort]{natbib}
\algnewcommand\algorithmicinput{\textbf{INPUT:}}
\algnewcommand\INPUT{\item[\algorithmicinput]}
\algnewcommand\algorithmicoutput{\textbf{OUTPUT:}}
\algnewcommand\OUTPUT{\item[\algorithmicoutput]}

\renewcommand{\selectlanguage}[1]{}
\newcommand{\mycomment}[1]{}
\newcommand{\myadd}[1]{{{#1}}}
\newcommand{\anantadd}[1]{{{#1}}}

\newcommand{\verbit}[1]{{\tt{#1}}}
\newcommand{\speakit}[1]{{/\tt{#1}/}}
\renewcommand{\L}{w_L}
\renewcommand{\P}{w_P}
\newcommand{\R}{\myadd{T}}
\newcommand{\E}{\myadd{T'}}
\newcommand{\mybox}[1]{\parbox[l]{0.75\textwidth}{\verbit{#1}}}
\newcommand{\K}{\R}
\newcommand{\GA}{{\sc{Ga}}}
\newcommand{\KU}{{\sc{Ku}}}
\newcommand{\KI}{{\sc{Ki}}}
\newcommand{\PS}{{\sc{Ps}}}
\newcommand{\UQ}{\R}
\newcommand{\myboxp}[1]{$\left \{ \mybox{#1} \right .$}
\newcommand{\myboxnp}[1]{\left \{ \mybox{#1} \right .}
\newcommand{\Ew}{\E}
\newcommand{\Kw}{\K}

\authorrunning{Anantaram, Sunil}
\titlerunning{Adapting general-purpose speech recognition engine for QA}
 \begin{document}

\title{Adapting general-purpose speech recognition engine output for
domain-specific natural language question answering}

\author{C. Anantaram \and Sunil Kumar Kopparapu} 
\institute{C. Anantaram \at TCS Innovation Labs - Delhi,
ASF Insignia, Gwal Pahari, Gurgaon, India \\
\email{c.anantaram@tcs.com}
\and Sunil Kumar Kopparapu \at TCS Innovation Labs - Mumbai,
Yantra Park, Thane (West)
\email{sunilkumar.kopparapu@tcs.com}
}
\maketitle

\begin{abstract}

Speech-based natural language question-answering interfaces to
enterprise systems are gaining a lot of attention. General-purpose
speech engines can be integrated with NLP systems to provide such
interfaces. Usually, general-purpose speech engines are trained on
large `general' corpus. However, when
such engines are used for specific domains, they may not recognize
domain-specific words well, and may produce erroneous output. Further,
the accent and the environmental conditions in which the speaker speaks
a sentence may induce the speech engine to inaccurately recognize
certain words. The subsequent natural language question-answering does
not produce the requisite results as the question does not accurately
represent what the speaker intended. Thus, the speech
engine's output may need to be adapted for a domain
before further natural language processing is carried out. We present
two mechanisms for such an adaptation, one based on evolutionary
development and the other based on machine learning, and show how we
can repair the speech-output to make the subsequent natural language
question-answering better.
\end{abstract}

\section{Introduction}
\label{sec:introduction}
Speech-enabled natural-language question-answering interfaces to
enterprise application systems, such as Incident-logging systems,
Customer-support systems, Marketing-opportunities systems, Sales data
systems etc., are designed to allow end-users to speak-out the
problems/questions that they encounter and get automatic responses. The
process of converting human spoken speech into text is performed by an
Automatic Speech Recognition (ASR) engine. While functional examples of
ASR with enterprise systems can be seen in day-to-day use, most of
these work under constraints of a limited domain, and/or use of
additional domain-specific cues to enhance the speech-to-text
conversion process. Prior speech-and-natural language interfaces for
such purposes have been rather restricted to either Interactive Voice
Recognition (IVR) technology, or have focused on building a very
specialized speech engine with domain specific terminology that
recognizes key-words in that domain through an extensively customized
language model, and trigger specific tasks in the enterprise
application system. This makes the interface extremely specialized,
rather cumbersome and non-adaptable for other domains. Further, every
time a new enterprise application requires a speech and natural
language interface, one has to redevelop the entire interface again.

An alternative to domain-specific \myadd{speech recognition}
engines has been to \mycomment{incorporate} \myadd{re-purpose}
general-purpose speech \myadd{recognition} engines, 
such as \mycomment{the speech engines in} Google \myadd{Speech API, IBM Watson
Speech to text API}
\mycomment{Now or Siri \citep{phonearena} \citep{acentral},}
which can be used across domains with natural language question
answering systems. Such general-purpose \myadd{automatic} speech engines
\myadd{(gp-ASR)} are \myadd{deep} 
trained on \myadd{very} large {\it{general}} corpus 
\myadd{using deep neural network (DNN) techniques. The deep learnt} 
\mycomment{incorporate}
acoustic and language models \myadd{enhance the performance of a ASR.} 
\mycomment{to cover such corpus.} However, this comes
with its own limitations. For freely spoken natural language sentences,
the typical recognition accuracy achievable even for state-of-the-art
speech recognition systems have been observed to be about $60$\% to $90$\%
in real-world environments \citep{lee_2010}. 
The recognition is worse
if we consider factors such as domain-specific words, environmental
noise, variations in accent, poor ability to express on the part of the
user, or inadequate \myadd{speech and language} resources from the domain to train such speech
recognition systems. The subsequent natural language processing, such
as that in a question answering system, of such erroneously and
partially recognized text becomes rather problematic, as the domain
terms may be inaccurately recognized or linguistic errors may creep
into the sentence. It is, hence, important to improve the accuracy of
the \myadd{ASR} output text. 

In this paper, we focus on the issues of using a readily available
\myadd{gp-ASR} \mycomment{general-purpose speech engine} and adapting its output for domain-specific
natural language question answering \citep{ijcai_2015}.
We present two
mechanisms for adaptation, namely 
\begin{enumerate}
\item[(a)] an evolutionary development based
artificial development mechanism of adaptation (Evo-Devo), where
we
consider the \myadd{output of }ASR \mycomment{output} 
as a biological entity that needs to adapt
itself to the environment (in this case the enterprise domain) through
\myadd{a mechanism of} repair and development of its
{\it{genes}}  and 
\item[(b)] a
machine learning based mechanism
where we examine the closest set of matches with trained
examples and the number of adaptive transformations that the ASR output
needs to undergo in order to be categorized as an acceptable natural
language input for question-answering. 
\end{enumerate}
We present the results of \mycomment{such}
\myadd{these two} adaptation and gauge the usefulness of each mechanism.
\myadd{The rest of the paper is organized as follows, in Section
\ref{sec:related_work} we briefly describe the work done in this area which
motivates our contribution. The main contribution of our work is
captured in Section \ref{sec:our_work} and we show the performance of our
approach through experiments in Section \ref{sec:experiments}. We conclude in
Section \ref{sec:conclusions}.}

\section{Related Work}
\label{sec:related_work}

Most work on ASR error detection and correction has focused on using
confidence measures, \myadd{generally called the log-likelihood score,} 
provided by the speech recognition engine\myadd{; the text with lower
confidence is assumed to be incorrect and subjected to correction.}
\mycomment{used to  and using
that to correct the ASR output.} Such confidence based methods are
useful only when we have access to the internals of a speech
recognition engine built for a specific domain. As mentioned earlier,
\myadd{use of} domain-specific engine requires one to rebuild the interface 
every time
the domain is updated, or a new domain is introduced. \myadd{As mentioned earlier, our
focus is to avoid rebuilding the interface each time the domain changes by}
\mycomment{In this paper, we focus on}
using an existing ASR. \myadd{As such} 
\mycomment{and thus} our method is
specifically a post-ASR system. A post-ASR system provides greater
flexibility in terms of \myadd{absorbing} \mycomment{modeling} domain variations and adapting the
output \myadd{of ASR} in ways that are not possible during training a
\myadd{domain-specific} ASR system
\citep{541124}.
\begin{quote} \em
\myadd{Note that an} \mycomment{An} 
erroneous ASR output text will lead to an equally (or more) erroneous
interpretation by the natural language question-answering system, resulting in
\myadd{a poor performance of the overall QA system} \mycomment{incorrect
output.}\mycomment{"}
\end{quote}

Machine learning classifiers have been used in the past for the purpose
of combining features to calculate a confidence score for error
detection. Non-linguistic and syntactic knowledge for detection of
errors in ASR output, using a support vector machine to combine
non-linguistic features was proposed in \citep{shi_2008} 
and Naive Bayes
classifier to combine confidence scores at a word and utterance level,
and differential scores of the alternative hypotheses was used in
\citep{1385606}
Both \citep{shi_2008} and \citep{1385606} 
rely on the
availability of confidence scores output by the ASR engine. A
syllable-based noisy channel model combined with higher level semantic
knowledge for post recognition error correction, independent of the
internal confidence measures of the ASR engine is described in
\citep{jeong_2004}.
\myadd{In \citep{Lopez-Cozar:2008:APS:1393642.1393764} the authors propose 
a method to correct errors in spoken dialogue systems. 
They consider several contexts to correct the speech recognition output 
including learning a threshold during training to decide when the correction 
must be carried out in the context of a dialogue system. They however use the confidence scores associated with the output text to do the correction or not.
The correction is carried using syntactic-semantic and lexical models 
to decide whether a recognition result is correct. 

In \citep{DBLP:journals/corr/abs-1203-5262} the authors proposes a 
method to detect and correct ASR output based on Microsoft N-Gram dataset.
They use a context-sensitive error correction algorithm for selecting the best candidate for correction using the Microsoft N-Gram dataset which contains real-world data and word sequences extracted from the web which can mimic a comprehensive dictionary of words having a large and all-inclusive vocabulary.

In \citep{6138180} the authors assume the availability of pronunciation primitive characters as the output of the ASR engine and then use 
domain-specific named entities to establish the context, 
leading to the correction of the  speech recognition output. The patent 
\citep{amento2007error} proposes a manual correction of the ASR output transcripts by providing visual display suggesting the correctness of the text output by ASR. Similarly, \citep{interspeech_2014} propose a 
re-ranking  and  classification  strategy based on  logistic regression model to estimate
the  probability  for  choosing  word  alternates 
to display to the user in their framework of a tap-to-correct
interface. 

}

Our proposed machine learning based system is along the lines of
\citep{jeong_2004} but with differences: (a) while they use a single feature
(syllable count) for training, we propose the use of multiple features
for training the Naive Bayes classifier, and (b) we do not perform any
manual alignment between the ASR and reference text -- this is done
using a\myadd{n edit distance} \mycomment{DTW-}based technique for sentence alignment. Except for
\citep{jeong_2004} all reported work in this area make use of features from
the internals of the ASR engine for ASR \myadd{text output} error detection. 

\mycomment{\centerline{\myadd{\em This section needs to be made richer.}}}

\myadd{We assume the use of a gp-ASR in the rest of the paper. Though we use
examples of natural language sentences in the form of queries or questions, it
should be noted that the description is applicable to any conversational 
natural language sentence.}

\section{Domain adaptation of ASR output}
\label{sec:our_work}

\subsection{Errors in \myadd{ASR} \mycomment{general-purpose ASR} output}
In this paper we focus on question answering interfaces to  enterprise
systems\myadd{, though our discussion is valid for any kind of natural language processing sentences that are not necessarily a query.} For example, suppose we have a retail-sales management system domain,
then end-users would be able to query the system through spoken natural
language questions ($S$) such as


$$S = \left \{ \mybox{{\speakit{What is the total sales
of miscellaneous store retailers from year two thousand ten to year two
thousand fifteen?}}} \right . $$ 
A perfect ASR would take $S$ as the input and produce ($T$), namely,
$$T = \left \{ \mybox{{{what is the total sales
of miscellaneous store retailers from year two thousand ten to year two
thousand fifteen}}} \right . $$ 
We consider the situation where
a ASR takes such a sentence ($S$) spoken by a person as
input, and outputs an inaccurately recognized text ($T'$) sentence. In our
experiments, when the above question was spoken by a person and processed by a popular ASR engine such as Google Speech API,
the output text sentence was ($T'$) 
$$T' = \left \{{\mybox{{{{what is the total
sales of miscellaneous storyteller from the year two thousand ten to
two thousand fifteen}}}}} \right .$$

\myadd{Namely $$S \longrightarrow \fbox{\mbox{ASR}} \longrightarrow T'$$} 
\myadd{It should be noted that an inaccurate output by the ASR engine maybe the result of various factors such as background noise, accent of the person speaking the sentence, the speed at which he or she is speaking the sentence, domain-specific words that are not part of popular vocabulary etc.} The subsequent natural
language question answering system cannot answer the above output sentence
from its retail sales data. Thus the question we tackle here is -- how
do we adapt or repair the sentence ($T'$) back to the original sentence ($T$) as
intended by the speaker.
\myadd{Namely $$T' \longrightarrow \fbox{\mbox{adaptation, repair}} \longrightarrow T$$} 
We present two mechanisms for adaptation or
repair of the ASR output, \myadd{namely $T' \longrightarrow T$,}  
in this paper: (a) an evolutionary development
based artificial development mechanism, and (b) a machine-learning
mechanism.


 \subsection{Evo-Devo based Artificial Development mechanism of adaption}

 \anantadd{
 Our mechanism is motivated by Evolutionary Development (Evo-Devo) processes in biology 
\citep{harding_2008,ca_icds_2014,tufte_2008} to help adapt/repair the overall content 
accuracy of an ASR output \myadd{($T'$)} for a domain. We give a very brief overview of Evo-Devo process in 
biological organisms and discuss how this motivates our mechanism. In a biological organism, 
evo-devo processes are activated when a new biological cell needs to be formed or an injured 
cell needs to be repaired/replaced. During such cell formation or repair, the basic genetic 
structure consisting of the genes of the organism are replicated into the cell -- the 
resultant set of 'genes in the cell' is called the {\it{genotype}} of the cell. Once this is 
done, the genotype of the cell is then specialized through various developmental processes 
to form the appropriate cell for the specific purpose that the cell is intended for, in 
order to factor-in the traits of the organism -- called the {\it{phenotype}} of the cell. 
For example, if a person has blue-eyes then a blue-eye cell is produced, or \myadd{if} a person has 
brown skin then a brown-skin cell is produced. During this process, environmental influence 
may also play a role in the cell's development and such influences are factored into the 
genotype-phenotype development process. The field of Evo-Devo has influenced the field of 
Artificial Intelligence  \myadd{(AI)} and a new sub-field called Artificial Development (Art-Dev) has been 
created that tries to apply Evo-Devo principles to find elegant solutions to adaptation and 
repair problems in AI.

We take inspiration from the Evo-Devo biological process and suitably tailor it to our 
research problem of repairing the ASR output \myadd{($T'$)}. In our approach we consider the erroneous ASR 
output text as the input for our method and treat it as an 'injured biological cell'. We 
repair that 'injured cell' through the development of the partial gene present in the input 
sentence with respect to the genes present in the domain. We assume that we have been 
provided with the domain ontology describing the terms and relationships of the domain. In 
our framework, we consider the domain ontology as the true 'genetic knowledge' of that 
'biological organism'. In such a scenario, the 'genetic repair' becomes a sequence of 
match-and-replace of words in the sentence with appropriate domain ontology terms and 
relationships. Once this is done, the 'genotype-to-phenotype repair' is the repair of 
linguistic errors in the sentence after the 'genetic repair'. The following sub-section 
describes our method in detail.

\subsubsection{{Repair method}}

We assume that all the instances of the objects in the domain are stored in a database 
associated with the enterprise system, and can be expressed in relational form (such as [a R 
c]), for example \verbit{[{\textquotesingle}INDUSTRY{\textquotesingle}, 
{\textquotesingle}has{\textquotesingle}, {\textquotesingle}PEAK SALES{\textquotesingle}]}. A 
relational database will store it as a set of tables and we treat the data in the database 
as static facts of the domain. The ontology of the domain can then be generated from this 
database. We assume that the data schema and the actual data in the enterprise application 
forms a part of the domain terms and their relationships in the ontology. This identifies 
the main concepts of the domain with a {\textless}subject- predicate-object{\textgreater} 
structure for each of the concepts. The ontology thus generated describes the relations 
between domain terms, for example \verbit{[{\textquotesingle}SALES{\textquotesingle}, 
{\textquotesingle}has\_code{\textquotesingle}, 
{\textquotesingle}NAICS\_CODE{\textquotesingle}]} or 
\verbit{[{\textquotesingle}OPTICAL\_GOODS{\textquotesingle}, 
{\textquotesingle}sales\_2009{\textquotesingle}, 
{\textquotesingle}8767\_million{\textquotesingle}]} and thus can be expressed using OWL 
schema as {\textless}s-p-o {\textgreater} structure. Each {\textless}s-p-o{\textgreater} 
entry forms the genes of the domain.

We start by finding matches between domain ontology terms and words that appear in the input 
sentence. Some words of the input sentence will match domain ontology terms exactly. The 
corresponding domain ontology entry consisting of subject-predicate-object triple is put 
into a candidate set. Next, other words in the input sentence that are not exact matches of 
domain ontology terms but have a 'closeness' match with terms in the ontology are 
considered. This 'closeness' match is performed through a mix of phonetic match combined 
with Levenshtein distance match. The terms that match help identify the corresponding domain 
ontology entry (with its subject-predicate-object triple) is added to the candidate set. 
This set of candidate genes is a shortlist of the 'genes' of the domain that is probably 
referred to in the input sentence.

Next, our mechanism evaluates the {`}fittest{'} domain ontology entry from the candidate set 
to replace the partial gene in the sentence. A fitness function is defined and evaluated for 
all the candidate genes short-listed. This is done for all words / phrases that appear in the 
input sentence except the noise words. The fittest genes replace the {\it{injured}} genes of 
the input sentence. The set of all genes in the sentence forms the {\it{genotypes}} of the 
sentence. This is the first-stage of adaptation.

Once the genotypes are identified, we grow them into {\it{phenotypes}} to remove the 
grammatical and linguistic errors in the sentence. To do this, we find parts of the sentence 
that is output by the first-stage of adaptation (the gene-level repair) and that violate 
well-known grammatical/ linguistic rules. The parts that violate are repaired through 
linguistic rules. This is the second stage of adaptation/ repair. This process of artificial 
rejuvenation improves the accuracy of the sentence, which can then be processed by a natural 
language question answering system \citep{Bhat:2007:FTC:1775431.1775465}. Thus, this 
bio-inspired novel procedure helps adapt/repair the erroneously recognized text output by a 
speech recognition engine, in order to make the output text suitable for deeper natural 
language processing. The detailed steps are described below.}
\begin{figure}[h]
\fbox{\vbox{
The fitness function $F$ takes as input the $asr\_word$, the candidate $gene$, the 
Levenshtein distance weight ($\L$), the Phonetic algorithm weight ($\P$) and Threshold 
($T$). Fitness function $F$ then tries to find the closeness of the match between 
$asr\_word$ and the candidate $gene$. To do that, the function calculates two scores: 
\begin{enumerate} \item $algoScore$: is an aggregated score of the similarity of the $gene$ 
with the $asr\_word$ by various phonetic algorithms; and \item $editScore$: is the 
Levenshtein distance between the $asr\_word$ and the $gene$. \end{enumerate} The fitness 
function then calculates the final fitness of the $gene$ using the formula: \begin{equation} 
finalScore = \P * algoScore + \L*(1-editScore) \label{eq:ed_cost} .\end{equation} If the 
$finalScore$ is greater than a given threshold $T$ the $asr\_word$ is replaced by the 
candidate $gene$, otherwise the $asr\_word$ is kept as it is, namely,
\[ if (finalScore > T)\;\;\; asr\_word \leftarrow gene \]
}}
\label{fig:fitness}
\caption{Fitness Function.}
\end{figure}
\vspace{2ex} \\
\noindent {{\bf Step 1: Genes Identification: }} 
We match the sub-parts (or sub-strings) of the ASR-output sentence with the genes of the 
domain. The match may be partial due to the error present in the sentence. The genes in the 
domain that match the closest, evaluated by a phonetic and/or syntactic match between the 
ontology entity and the selected sub-part, are picked up and form the candidates set for the 
input sentence. For example, let the actual sentence that is spoken by an end-user be 
\verbit{{"}which industry has the peak sales in nineteen ninety seven?{"}}. In one of our 
experiments, when Google Speech API was used as the ASR engine for the above sentence spoken by a 
user, then the speech engine{'}s output sentence was \verbit{{"}which industry has 
the pixel in nineteen ninety seven?{"}}. This ASR output is erroneous (probably due to 
background noise or the accent of the speaker) and needs repair/ adaptation for the domain.

As a first step, the ASR output sentence is parsed and the Nouns and Verbs are identified 
from part-of-speech (POS) tags. Syntactic parsing also helps get {\textless}subject-verb-object{\textgreater} 
relations to help identify a potential set of 
{\textless}s-p-o{\textgreater} genes from the ontology. For each of the 
Nouns and Verbs and other syntactic relations, the partially matching genes with respect to 
the domain ontology are identified; for this particular sentence the partially matching 
genes are, \verbit{{"}industry{"}} and \verbit{{"}pixel{"}}. This leads us to identify the 
probable set of genes in the domain ontology that are most likely a possible match: 
\verbit{{\textquotesingle}INDUSTRY{\textquotesingle}, 
{\textquotesingle}has{\textquotesingle}, {\textquotesingle}PEAK SALES{\textquotesingle}}. 
The set of all such probable genes need to be evaluated and developed further.
\vspace{2ex} \\
\noindent {\bf Step 2: Developing the genes to identify the genotypes:}
Once the basic candidate genes are identified, we evaluate the genes to find the best fit 
for the situation on hand with evolution and development of the genes, and then test a 
fitness function (see Fig. \ref{fig:fitness} and select the most probable gene that survives. This gives us the set of 
genotypes that will form the correct ASR sentence. For example, the basic genes 
\verbit{{"}INDUSTRY{"}} and \verbit{{"}PIXEL{"}} are used to match the substring 
\verbit{{"}industry has the pixel{"}} with the gene \verbit{{"}INDUSTRY{\textquotesingle}, 
{\textquotesingle}has\_field{\textquotesingle}, {\textquotesingle}PEAK SALES{'}}.  This is 
done through a matching and fitness function that would identify the most appropriate gene 
of the domain. We use a phonetic match function like Soundex, Metaphone, Double-metaphone 
\citep{naumann_2015} to match \verbit{{"}pixel{"}} with \verbit{{"}PEAK SALES{"}} or an 
edit-distance match function like Levenshtein distance \citep{naumann_2015} to find the 
closeness of the match. In a large domain there may be many such probable candidates. In 
such a case, a fitness function is used to decide which of the matches are most suitable. 
The genes identified are now collated together to repair the input sentence. This is done by 
replacing parts of the input sentence by the genes identified in the previous step. In the 
above example the ASR sentence, \verbit{{"}Which industry has the pixel in nineteen ninety 
seven?{"}} would be adapted/repaired to \verbit{{"}Which industry has the peak sales in 
nineteen ninety seven?{"}}.
\vspace{2ex} \\
\noindent {\bf Step 3: Developing Genotypes to Phenotype of sentence:}
The repaired sentence may need further linguistic adaptation/ repair to remove the remaining 
errors in the sentence. To achieve this, the repaired ASR sentence is re-parsed and the POS 
tags are evaluated to find any linguistic inconsistencies, and the inconsistencies are then 
removed. For example, we may notice that there is a WP tag in a sentence that refers to a 
Wh-Pronoun, but a WDT tag is missing in the sentence that should provide the Determiner for 
the Wh-pronoun. Using such clues we can look for phonetically matching words in the sentence 
that could possibly match with a Determiner and repair the sentence. Linguistic repairs such 
as these form the genotype to phenotype repair/ adaptation of the sentence. The repaired 
sentence can then be processed for question-answering.

We use open source tools like LanguageTool to correct grammatical errors. 
In addition we have added some domain specific grammar rules.
As we understand, the LanguageTool has $54$ grammar rules, $5$ style rules and $4$ built-in Python rules for grammar check and correction.
Further we have added some $10$ domain specific rules to our linguistic repair function. Our grammar rules can be extended or modified for any domain.

\subsubsection{{Algorithm of the Evo-Devo process}}

The algorithm has two main functions: ONTOLOGY\_BASED\_REPAIR (that encode Steps $1$ and $2$ 
described above) and LINGUISTIC\_REPAIR (encoding Step $3$ above). The input sentence 
is POS\_tagged and the nouns and verbs are considered. 
A sliding window allows the algorithm to consider single words or 
multiple words in a domain term.

Let $S = {w_1, w_2, w_3, \cdots,  w_n}$ be the set of words in the 
ASR-output(asr\_out). 
Let $D= {dt_1, dt_2, dt_3, \cdots, dt_m }$ be the domain-ontology-terms. 
These terms may be considered as candidate genes that can possibly replace the 
ASR output (asr\_out) words that may be erroneously recognized. 
A sliding window of length $l$ consisting of words ${w_i, \cdots, w_{i+l-1}}$ is 
considered for matching with domain-ontology-terms. 
The length $l$ may vary from $1$ to $p$, where $p$ may be decided based on the 
environmental information. For example, if the domain under consideration 
has financial terms then $p$ may be five words, while for a domain pertaining to car parts, 
$p$ may be two words.
\newcommand{\ewi}{\phi}
The part\_match functionality described below evaluates a cost function, say
$ C(\{w_i,\cdots, w_{i+l-1}\}, dt_k)$ such that minimizing $C(\{w_i,\cdots, w_{i+l-1}\}, dt_k)$ 
would result in $dt_*$ which may be a possible candidate to replace $\{w_i,\cdots, w_{i+l-1}\}$, namely,
\[ dt_* = \min_{G} C(\{w_i,\cdots, w_{i+l-1}\}, dt_k) \]
The cost function 
\[C(\{w_i,\cdots, w_{i+l-1}\}, dt_k) = \]
\begin{eqnarray}
b_1 &*& \mbox{soundex}(\ewi\{w_i,\cdots, w_{i+l-1}\}, \ewi\{dt_k\}) + \nonumber \\
b_2 &*& \mbox{metaphone}(\ewi\{w_i,\cdots, w_{i+l-1}\}, \ewi\{dt_k\}) + \nonumber \\
b_3 &*& (\mbox{edit\_distance}(\ewi\{w_i,\cdots, w_{i+l-1}\}, \ewi\{dt_k\}) + \nonumber   \\
b_4 &*& (\mbox{number\_of\_syllables}(\ewi\{w_i,\cdots, w_{i+l-1}\}) - 
\mbox{number\_of\_syllables}(\ewi\{dt_k\} ) + \nonumber   \\
b_5&*& (\mbox{word2vec}(\ewi\{w_i,\cdots, w_{i+l-1}\}) - \mbox{word2vec}(\ewi\{dt_k\})^2 
\nonumber \end{eqnarray}
where weights, $b_1+b_2+b_3+b_4+b_5 = 1$ and $\ewi\{k\}$ represents each-element-in the set $k$.
If the value of the cost function $C(\{w_i,\cdots, w_{i+l-1}\}, dt_k)$ is greater than the 
pre-determined threshold $T$ then $\{w_i,\cdots, w_{i+l-1}\}$ may be replaced with the $dt_*$, 
otherwise the $\{w_i,\cdots, w_{i+l-1}\}$ is maintained as it is. 

The broad algorithm of Evolutionary Development mechanism is shown in
Algorithm \ref{algo:Evo-Devo-Mechanism}.

\begin{algorithm*}
\caption{Evo-Devo Mechanism}
\label{algo:Evo-Devo-Mechanism}
\begin{algorithmic}[1]
\INPUT ASR output sentence, \emph{sentence}; domain\_ontology
\OUTPUT Repaired sentence, \emph{repaired\_sentence}
\Statex start
\Comment{parse the input sentence}
\State parsed\_sentence $\leftarrow$ POS\_tag($sentence$)
\Comment{repair process starts - do genetic repair and find the genotypes}
\State part\_repaired\_sentence $\leftarrow$
\Call{ontology\_based\_repair}{parsed\_sentence}
\Comment{grow the genotypes into phenotypes}
\State repaired\_sentence $\leftarrow$
\Call{linguistic\_repair}{parsed\_sentence,part\_repaired\_sentence}
\Statex end
\Statex
\Function{ontology\_based\_repair}{parsed\_sentence}
\State nouns\_verbs $\leftarrow$ find(parsed\_sentence, noun\_verb\_POStoken)
\Comment for each noun\_verb\_entry in nouns\_verbs do next 4 steps
\Comment{find partially matching genes: match nouns and verbs with entries in domain ontology with phonetic algorithms and Levenshtein distance match}
\State concepts\_referred $\leftarrow$ part\_match(noun\_verb\_entry,
domain\_ontology)
\Comment{find genes: get the subject-predicate-object for concepts}
\State candidate\_genes $\leftarrow$ add(spo\_entry, concepts\_referred)
\Comment{simulate the development process of the genes - find the fittest gene from candidate genes}
\State fit\_gene $\leftarrow$ fittest(candidate\_genes, POS\_token)
\Comment{add fittest gene into set of genotypes}
\State genotypes $\leftarrow$ add(fit\_gene)
\Comment{replace partially identified genes in input with genotypes identified}
\State repaired\_sentence $\leftarrow$ substitute(parsed\_sentence, nouns\_verbs,
genotypes)
\State \Return repaired\_sentence
\EndFunction
\Statex
\Function{linguistic\_repair}{part\_repaired\_sentence}
\State other\_POStags $\leftarrow$ find(part\_repaired\_sentence, remaining\_POStokens)
\Comment{find POS tags without linguistic completion}
\State ling\_err $\leftarrow$ linguistic\_check( other\_POStags,
part\_repaired\_sentence)
\Comment{find candidate words for linguistic error}
\State candidate\_words $\leftarrow$ add(part\_repaired\_sentence, ling\_err)
\Comment{find the closest semantic match for error words}
\State fit\_word $\leftarrow$ fittest\_word(candidate\_words, ling\_err)
\Comment{add fittest word into repaired sentence}
\State fit\_words $\leftarrow$ add(candidate\_word, fit\_word)
\Comment{create the repaired sentence}
\State repaired\_sentence $\leftarrow$ replace(part\_repaired\_sentence,
fit\_words, other\_POStags)
\State \Return repaired\_sentence
\EndFunction
\end{algorithmic}
\end{algorithm*}

\subsubsection{{Detailed example of our method}}
Let us assume that we have the domain of retail sales data described in
an ontology of {\textless}subject-predicate-object{\textgreater}
structure as shown in Table \ref{fig:ont}.

\begin{table}[h]
\begin{center}
\begin{small}
\begin{tabular}{l|c|l} \hline
Subject & Predicate & Object \\ \hline
\verbit{INDUSTRY}    & \verbit{BUSINESS} &    \verbit{CAR\_DEALERS}  \\
\verbit{INDUSTRY}    & \verbit{BUSINESS}  &   \verbit{OPTICAL\_GOODS} \\
\verbit{CAR\_DEALERS} &      \verbit{SALES\_2013} &     \verbit{737640\_million}  \\
\verbit{CAR\_DEALERS} &      \verbit{SALES\_2011}  &    \verbit{610747\_million} \\
\verbit{CAR\_DEALERS} &      \verbit{SALES\_2009}    &  \verbit{486896\_million} \\
\verbit{OPTICAL\_GOODS}&     \verbit{SALES\_2013}    &  \verbit{10364\_million} \\
\verbit{OPTICAL\_GOODS}&     \verbit{SALES\_2011}    &  \verbit{10056\_million} \\
\verbit{OPTICAL\_GOODS}&     \verbit{SALES\_2009}    &  \verbit{8767\_million} \\ \hline
\end{tabular}
\end{small}
\end{center}
\caption{Ontology Structure.}
\label{fig:ont}
\end{table}

Now, let us consider that a user speaks the following sentence to Google
Now speech engine: \verbit{{"}Which business has more sales in
2013: Car dealers or optical goods?{"}}. In our
experiment the Google Now speech engine produced the ASR output
sentence as \verbit{{"}which business has more sales in 2013
car dealers for optical quotes{"}}. The recognized ASR
sentence has errors. In order to make this ASR sentence more accurate,
we input this sentence into the Evo-Devo mechanism and run the process:

\begin{itemize}
\item {\bf Genes Identification (Step 1):}
We parse the ASR sentence and identify the parts-of-speech in it as:
\verbit{which/WDT}, 
\verbit{business/NN}, 
\verbit{has/VBZ}, 
\verbit{more/JJR}, 
\verbit{sales/NNS}, 
\verbit{in/IN}, 
\verbit{2013/CD},
\verbit{car/NN}, 
\verbit{dealers/NNS}, 
\verbit{for/IN}, 
\verbit{optical/JJ}, 
\verbit{quotes/NNS}.

 Considering the words that have POS tags of Nouns (\verbit{NN/NNS} etc.) in the
example sentence we get the words
\verbit{{"}business{"}},
\verbit{{"}sales{"}},
\verbit{{"}car{"}},
\verbit{{"}dealers{"}},
\verbit{{"}quotes{"}}. Based on these words we
extract all the partially matching subject-predicate-object instances
of the domain ontology. For example, we obtain instances such as
\verbit{[OPTICAL\_GOODS}  \verbit{SALES\_2013}  \verbit{10364\_million]}, 
\verbit{[INDUSTRY}  \verbit{BUSINESS} \verbit{OPTICAL\_GOODS]}  and 
\verbit{[INDUSTRY  BUSINESS  CAR\_DEALERS]}, 
etc.
from the domain ontology that are partially matching with the words
\verbit{{"}business{"}} and
\verbit{{"}sales{"}} respectively. POS tag
\verbit{2013/CD} also leads to reinforcing the above {\textless}s-p-o{\textgreater} instance.

\item {\bf Developing the genes to identify the genotypes (Step 2):}
We replace the erroneous words in the sentence by using a fitness
function. The fitness function is defined using string similarity
metric (Levenshtein distance) and an aggregated score of phonetic
algorithms such as Soundex, Metaphone and Double Metaphone as described
in Fitness function in the section above. Thus we get the following
adaptation: \verbit{which business has more sales in 2013 car dealers for
optical goods?}

\item {\bf Developing Genotypes to Phenotype (Step 3):}
We now find the parts-of-speech of the repaired sentence after the step
2 as: \verbit{which/WDT}, 
\verbit{business/NN}, 
\verbit{has/VBZ}, 
\verbit{more/JJR}, 
\verbit{sales/NNS}, 
\verbit{in/IN},
\verbit{2013/CD}, 
\verbit{car/NN}, 
\verbit{dealers/NNS}, 
\verbit{for/IN}, 
\verbit{optical/JJ}, 
\verbit{goods/NNS}.

In the linguistic repair step, we find that since there is no direct
ontological relationship between \verbit{{"}car
dealers{"}} and \verbit{{"}optical
goods{"}}, we cannot have the preposition \verbit{for} between
these domain terms. Thus we have to find a linguistic relation that is
permissible between these domain terms. One of the options is to
consider linguistic relations like
{`}conjunction{'},
{`}disjunction{'} between domain terms.
Thus, when we evaluate linguistic relations AND or OR between these
domain terms, we find that OR matches closely with for through a
phonetic match rather than AND. Thus we replace \verbit{for} with \verbit{or} in the
sentence. Hence the final output of the Evo-Devo mechanism is
\verbit{{"}which business has more sales in 2013 car dealers or
optical goods?{"}}. This sentence can now be processed
by a question-answering (QA) system. In the above example, a QA system
\citep{Bhat:2007:FTC:1775431.1775465} would parse the sentence, identify the known
ontological terms \verbit{\{business, sales, 2013, car dealers, optical
goods\}}, find the unknown predicates \{which business, more sales\},
form the appropriate query over the ontology, and return the answer
\verbit{{"}CAR\_DEALERS{"}}.

\end{itemize}

\anantadd{
\subsubsection{{Limitations of the method}}

We assume that there is a well-structured domain ontology for the domain and it is  available in the form of {\textless}s-p-o{\textgreater} triples.
We also assume that the speaker speaks mostly grammatically correct sentences using terms in the domain. While the method would work for grammatically incorrect sentences, 
the linguistic repair step would suffer.

We assume that the speech is processed by a gp-ASR and the ASR-output forms the 
input sentence that needs repair.
However,
it is important to note that the input sentence (i.e. the ASR output) need
not necessarily contain {\textless}s-p-o{\textgreater} triples for our method to work. The 
{\textless}s-p-o{\textgreater} triples that are short-listed from domain ontology aid in  
forming a candidate set of 'possible genes' to consider and the fittest amongst them is 
considered (Step 2) in the context of the other words in the sentence.
For example, if the input sentence was \verbit{'Who had pick sales'} 
would get repaired to \verbit{'Who had peak sales'} since the domain term of 
\verbit{'peak sales'} would match with
\verbit{'pick sales'} in our method.
Further, input sentences need not necessarily be queries; 
these can be just statements about a domain.
For example, if the above ASR-output sentence was 
\verbit{"We hit pick sales this season"}, the method would repair it as  
\verbit{"We hit peak sales this season"} using the same set of steps for repair.
However, as of now, our method does not repair paraphrases of sentences like 
\verbit{"which industry had biggest sales"} to \verbit{"which industry had peak sales"}. 
Such repairs need extension to our matching process.

The method does not impose any restriction on the sentence or its formation; 
it can be a fully meaningful sentence in a domain or may contain partial information. 
The method finds the fittest repair for the inaccuracies occurring in an sentence, 
post-ASR recognition. It should also be noted that the method does not know the 
original sentence spoken by the speaker, but tries to get back the original 
sentence for a particular domain.
}

\subsection{Machine Learning mechanism of adaptation}

In the machine learning \myadd{based} mechanism of adaptation, we assume the
availability of example pairs of \myadd{$(T', T)$ namely (ASR output, the
actual transcription of the spoken sentence)} for training.
\myadd{We further assume that} such a machine-learnt model can help repair an 
unseen ASR output to its
intended \myadd{correct} sentence. We address \myadd{the following hypothesis}
\mycomment{question:} \begin{quote} \em Using the information from
past recorded errors and the corresponding correction, can we learn how
to repair (and thus adapt to a new domain) the text after ASR? \end{quote} 
\myadd{Note that this is equivalent to, \myadd{albiet loosely,} 
learning the {\em error model} of a
 specific ASR.}
Since we
have a small training set, we have used the Naive Bayes classifier that
is known to perform well for small datasets with high bias and low
variance. We have used the NLTK \citep{Bird:2009:NLP:1717171} Naive Bayes
classifier in all our experiments. 

Let $\E$ be the erroneous text (which is the ASR output), $\R$ the
corresponding reference text (which is the textual representation of
the spoken sentence) and $F$ a feature extractor, \myadd{ such that 
\begin{equation}
f_\beta = F(\E_\beta)
\label{eq:f_extraction}
\end{equation}
where \begin{equation} f_\beta = 
( f_{\beta 1}, 
f_{\beta 2},  \cdots
f_{\beta n} ) 
\label{eq:feature_expand}
\end{equation} is a set of $n$ features extracted from $\E_\beta$.
}
Suppose there
are several pairs say ($\E_i$, $\R_i$) for $i=1, 2, {\cdots} N$. 
\myadd{Then we can derive $f_i$ 
for each $\E_i$ using (\ref{eq:f_extraction})}. 
The probability that $\E_k$ belongs to the
class $\K_k$ can be derived through the feature set $f_k$ as
follows.
\[
P(\K_k | f_k) = \frac{P(\K_k) * P(f_k | \K_k)}{P(f_k)}
\]
where $P(\K_k)$ is the apriori probability of the class $\K_k$ \myadd{and}
$P(f_k | \K_k)$ is the probability of occurrence of the
features $f_k$ in the class $\K_k$, and $P(f_k)$ is the overall
probability of the occurrence of the feature set $f_k$. Making naive 
assumption of independence in the features $f_{k1}, f_{k2}, {\cdots}
f_{kn}$ we get

\begin{equation}
P(\K_k| f_k)=\frac{P(\K_k)*(P(f_{k1} | 
\K_k)*P(f_{k2} | \K_k)*{\cdots}
P(f_{kn} | \K_k))}{ P(f_k)} 
\label{eq:eqn2}
\end{equation}
In our experiments, the domain specific reference text
$\K_i$ was spoken by several people and the spoken
speech was passed through a general purpose speech recognition engine
(\myadd{ASR}) that produced a (possibly) erroneous hypothesis
$\E_i$. Each pair of reference and the ASR output
(i.e. hypothesis) was then word aligned using edit distance, and the
mismatching pairs of words were extracted as $(\E_i, \K_i)$ pairs.
For example, if we have the following spoken sentence:
$$S_1: \left \{ \mybox{\speakit{In which year beer wine and liquor stores has
successful year}} \right . $$ 
and the corresponding true transcription
$$\R_1: \left \{ \mybox{{In which year beer wine and liquor stores has
successful year}} \right . $$ 
One of the corresponding ASR output $\E_1$ was
$$\E_1:  \left \{ \mybox{in which year dear wine and liquor stores have
successful year} \right . $$

In this case the $(\Ew, \Kw)$ pairs are \verbit{(dear, beer)} and \verbit{(have, has)}. As
another example consider that $\K_2$ was spoken but
$\E_2$ was recognized by the ASR.
$$\K_2: \left \{ \mybox{Whether the sales of jewelry business crosses
fifty thousand in a year } \right . $$
$$\E_2: \left \{ \mybox{whether the sales of than twenty business
crosses fifty thousand in a year } \right . $$
Clearly, in this case the $(\E, \K)$ pair is \verbit{(than twenty, jewelry)}.

Let us assume two features, namely, $f_\beta$ in (\ref{eq:f_extraction}) is of
dimension $n = 2$. Let the two features be  $(\mbox{number of words}, 
\mbox{number of syllables})$. Then, for \myadd{the  $(\E, \K)$ pair} \verbit{(than twenty, jewelry)} we have 
\[F(\verbit{(than\ twenty)}) = (2,3) \]
since the number of words in
\verbit{than twenty} is $2$ and \verbit{than twenty} contains
$3$ syllables. $P(f_{k1} | \K_k)$ in this case would be the
probability that the number of words in the input are two ($f_{k1}=2$)
when the
correction is \verbit{jewelry}. A third example is:
$$\K_3: \left \{ \mybox{In two thousand thirteen which industry had
the peak sales} \right . $$
$$\E_3: \left \{ \mybox{in two thousand thirteen which industry have
the pixels} \right . $$
Note that in this case the $(\E, \K)$ pair is \verbit{(peak sales, pixel)}.

Calculating thus the values of $P(\K_k)$ for all reference corrections,
$P(f_{kj} | \K_k)$ for all feature values, $f_{kj}$ for all
the $j$ features in $f_k$, we are in a position to calculate the RHS of
(\ref{eq:eqn2}). When this trained classifier is given an erroneous text, features
are extracted from this text and the repair works by replacing the
erroneous word by a correction that maximizes (\ref{eq:eqn2}), 
\[
\K_k^* = \max_{\K_k} P(\K_k | f_k)
\]
Namely, the $\K_k^*$ for which $P(\K_k | f_k)$ is maximum.

\section{Experiments and results}
\label{sec:experiments}

We present the results of our experiments with both the Evo-Devo and
the Machine Learning mechanisms \myadd{described earlier} using the 
U.S. Census Bureau conducted
Annual Retail Trade Survey of U.S. Retail and Food Services Firms for
the period of $1992$ to $2013$ \citep{retail}. 

\subsection{\myadd{Data Preparation}}
We downloaded this survey
data and hand crafted a total of $293$ textual questions \citep{awaz_yp} 
which could answer the survey data.
\myadd{A set of $6$ people (L2 English) generated $50$ queries each with the only constraint that these queries should be able to answer the survey data. 
In all a set of $300$ queries were crafted of which duplicate queries were removed to leave $293$ queries in all.}
Of these, we chose $250$
queries randomly and distributed among $5$ Indian speakers, who were
asked to read aloud the queries into a custom-built audio data
collecting application. So, in all we had access to $250$ audio queries
spoken by $5$ different Indian speakers; each speaking $50$ queries. 

\myadd{
\begin{figure}
\centering
\includegraphics[width=0.90\textwidth]{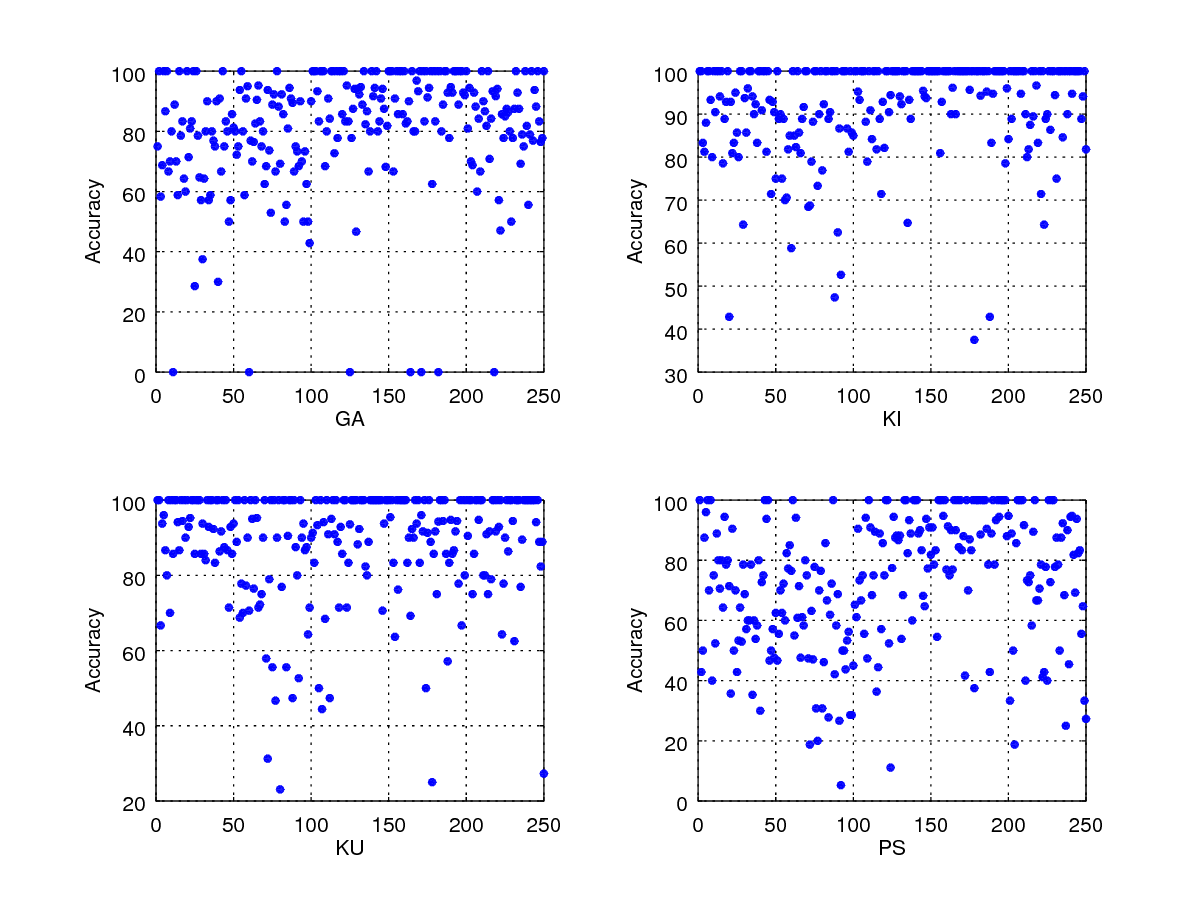}
\caption{$\E$ accuracy ($y$-axis) for the $250$ utterance ($x$-axis) for \GA, \KI, \KU\ and \PS.}
\label{fig:asr4_plot}
\end{figure}
}

Each of these $250$ audio utterances were
passed through $4$ different ASR engines, namely,  
Google ASR (\GA),  
Kaldi with US acoustic models (\KU),  
Kaldi with Indian Acoustic models  (\KI) and 
PocketSphinx ASR (\PS). 
\myadd{
In particular, that audio utterances were in wave format ({\tt .wav}) with a sampling rate of $8$ kHz and $16$ bit. 
In case of Google ASR (\GA), each utterance was first converted into {\tt .flac} format using the utility 
sound exchange ({\tt sox}) commonly available on {\tt Unix} machines. 
The {\tt .flac} audio files were sent to the cloud based Google ASR (\GA) one by one in a batch mode 
and the text string returned by \GA\ was stored. In all $7$ utterances did not get any text output, presumably \GA\ was unable to recognize the utterance. For all the other $243$ utterances a text output was received.

In case of the other ASR engines, namely, Kaldi with US acoustic models (\KU),
Kaldi with Indian Acoustic models  (\KI) and
PocketSphinx ASR (\PS) we first took the queries corresponding to the $250$ utterances and built 
a statistical language model (SLM) and a lexicon using the scripts that are available with PocketSphinx 
\citep{ps-lm} 
and Kaldi \citep{kaldi-lm}. This language model and lexicon was used with the acoustic model that were readily available with Kaldi and \PS. In case of \KU\  we used the American English acoustic models, while in case of \KI\ we used the Indian English acoustic model. In case of \PS\ we used the Voxforge acoustic models \citep{ps-am}.
Each utterance was passed through Kaldi ASR for two different acoustic models to get $\E$ corresponding to \KU\ and \KI. Similarly all the $250$ audio utterance were passed through the \PS\ ASR to get the corresponding $\E$ for \PS.
A sample utterance and the output of the four engines is shown in Figure \ref{fig:sample_out}.
\begin{figure}
{
$$S: \myboxnp{\speakit{Which stores has total sales more than two hundred thousand}} $$
$$\E:\myboxnp{which state has total sales more than twenty thousand}         \mbox{\GA; 70.00\%}$$
$$\E:\myboxnp{which stores has total sales more than two in two thousand}          \mbox{\KI; 80.00\%}$$
$$\E:\myboxnp{ which stores has total sales more than point of sales}          \mbox{\KU; 70.00\%}$$
$$\E:\myboxnp{ list the total sales more than}          \mbox{\PS; 40.00\%}$$
}
\caption{Sample output ($\E$) of four different ASR for the same spoken utterance ($S$). Also shown are the accuracy of the ASR output.}
\label{fig:sample_out}
\end{figure}
}

\myadd{Figure \ref{fig:asr4_plot} and Table \ref{tab:2} capture the performance of the different speech recognition engines.}
The performance of the ASR engines varied, with
\KI\ performing the best with $127$ of the $250$ utterances being 
correctly recognized
while \PS\ returned only $44$ correctly recognized utterances (see Table
\ref{tab:2}, Column $4$ named "Correct") of $250$ utterances.  
The accuracy of the ASR varied widely. For instance, in
case of \PS\ there were as many as $97$ instances of the $206$ erroneously 
recognized utterances which had an accuracy of
less than $70$\%. 

\myadd{
\begin{figure}
\centering
\includegraphics[width=0.90\textwidth]{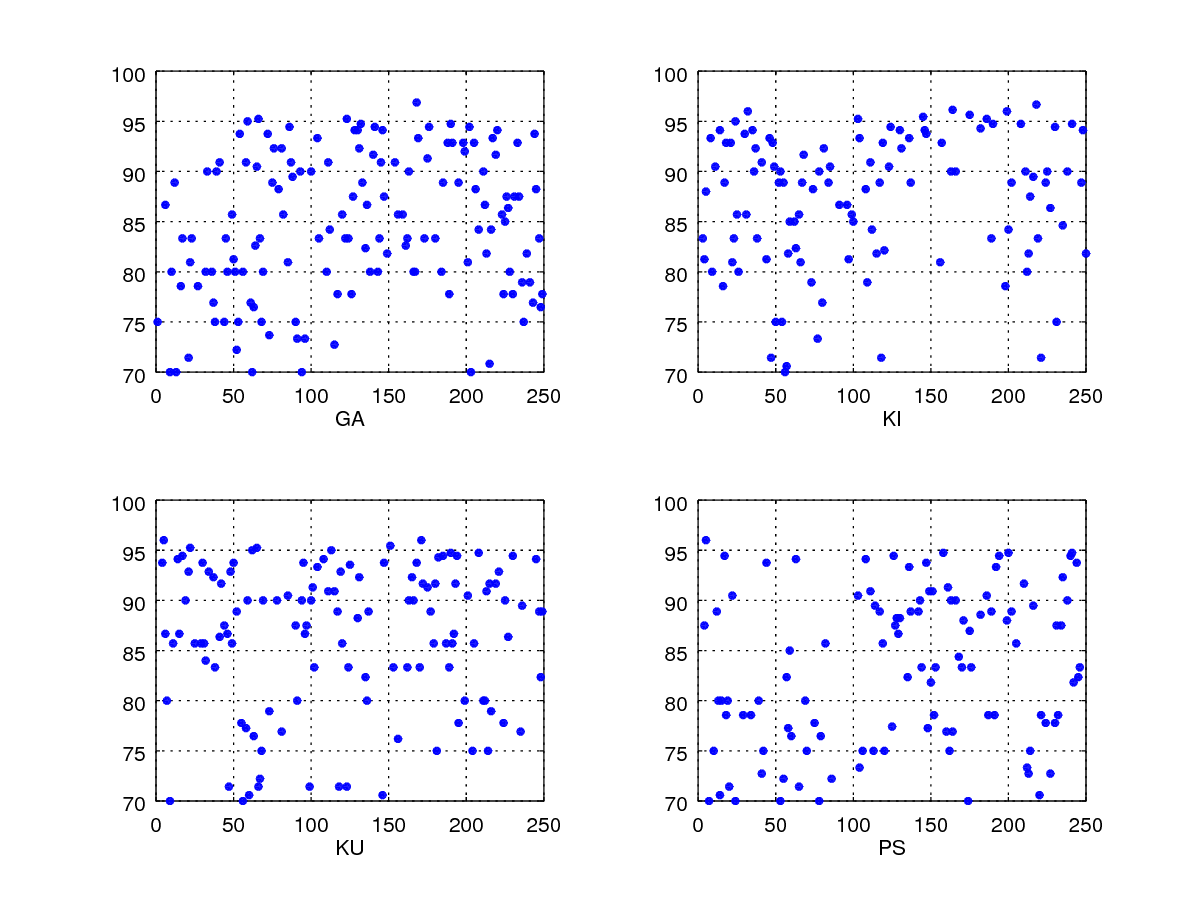}
\caption{All utterances that have and $\E$ accuracy ($y$-axis) $\ge 70$ and $<100$ used in all our experiments.}
\label{fig:asr4_used}
\end{figure}
}

Note that the accuracy is computed as the number of
deletions, insertions, substitutions that are required to convert the
ASR output to the textual reference (namely, $\E \rightarrow \R$) and is a common metric used in
speech literature \citep{HUNT1990329}. 

For all our analysis, we used only those utterances
that had an accuracy $70$\% but less that $100\%$, namely, $486$ instances (see Table
\ref{tab:2}, Figure \ref{fig:asr4_used}).
\myadd{An example showing the same utterance being recognized by four different ASR engines
is shown in Figure \ref{fig:sample_out}. 
Note that we used $\E$ corresponding to \GA, \KI\ and \KU\ in our analysis (accuracy $\ge 70\%$) and not $\E$ corresponding to \PS\
which has an accuracy of $40\%$ only. This is based on our observation that any ASR output that is 
lower that $70\%$ accurate is so erroneous that it is not possible to adapt and steer it towards the expected output. 
}
\begin{table}
\begin{center}
\begin{tabular}{|l||c|c||l|c|c|c|} \hline
ASR engine & Result & No result & Correct & Error & {\textgreater}=70\% & {\textless}70\% \\
& & & & (A+B) & (A)& (B) \\\hline
Google ASR  (\GA)& 243 & 7 & 55 & 188 & 143 & 45\\\hline
Kaldi US  (\KU)& 250 & 0 & 103 & 147 & 123 & 24\\\hline
Kaldi IN  (\KI)& 250 & 0 & 127 & 123 & 111 & 12\\\hline
PocketSphinx (\PS) & 250 & 0 & 44 & 206 & 109 & 97\\\hline\hline
Total & 993 & 7 & 329 & 664 & 486 & 178\\\hline
\end{tabular}
\end{center}
\caption{ASR engines and their output \%accuracy}
\label{tab:2}
\end{table}
The ASR output ($\E$)  are then given as input in the Evo-Devo and Machine
Learning mechanism of adaptation.

\subsection{ Evo-Devo based experiments}

We ran our Evo-Devo mechanism with the \mycomment{$250$} \myadd{$486$} ASR sentences \myadd{(see Table \ref{tab:2})}
and measured the accuracy after each repair. On an
average we have achieved about $5$ to $10$\% improvements 
in the \myadd{accuracy of the} sentences.
Fine-tuning the repair and fitness functions\myadd{, namely Equation 
(\ref{eq:ed_cost}),}  would probably yield much
better performance accuracies. However, experimental results confirm
that the \myadd{proposed} Evo-Devo mechanism is an approach that is able to adapt $\E$ to get closer to $\R$. We present a
snapshot of the experiments with Google ASR (\GA)
and calculate
accuracy with respect to the user spoken question as shown in 
Table \ref{tab:3}.

\begin{table*}
\begin{tabularx}{1.03\textwidth}{|X|c|} \hline
User's Question ($\UQ$), Google ASR out ($\E_{\mbox{\GA}}$), After Evo-devo ($\E_{ED}$) &
Acc \\ \hline
$\UQ$: \myboxp{In two thousand fourteen which industry had the peak sales} & \\ 
$\E_{\mbox{\GA}}$: \myboxp{in two
thousand fourteen which industry had the pixels} &  \GA: 80\%\\
$\E_{ED}$:  \myboxp{in two thousand fourteen
which industry had the peak sales} & ED: 100\% \\ \hline\hline
$\UQ: \myboxnp{in which year did direct selling establishments make the maximum sales and
in which year did they do the minimum sales}$ & \\
$\E_{\mbox{\GA}}: \myboxnp{which year did direct selling
establishments make the maximum cells and in which year did they do the many
muscles}$ & \GA:80.9\%\\
$\E_{ED}: \myboxnp{which year did direct selling establishments make the maximum sales
and in which year did they do the many musical}$  &   ED: 85.7\%
�\\ \hline\hline
$\UQ: \myboxnp{Which one among the  electronics and  appliance store and food and
beverage stores  has sales in more than hundred thousand in at least three
years in a row}$ & \\ 
$\E_{\mbox{\GA}}: \myboxnp{which one among the electronics and appliance store and food
and beverages stores have sales in more than one lakh in at least three years
in a row}$ &  \GA: 85.7\%\\
$\E_{ED}$: \myboxp{which one among the electronics and appliance store and food and
beverage stores have sales in more than one lakh in at least three years in a
row}  &  ED:89.3\% \\ \hline\hline
\end{tabularx}
\caption{Evo-Devo experiments with Google ASR (\GA).}
\label{tab:3}
\end{table*}

Table \ref{tab:3} clearly demonstrates the promise of the evo-devo mechanism for
adaptation/repair. In our experiments we observed that the
adaptation/repair of sub-parts in ASR-output ($\E$) that most probably
referred to domain terms occurred well and were easily repaired, thus
contributing to increase in accuracy. For non-domain-specific
linguistic terms the method requires one to build very good linguistic
repair rules, without which the method could lead to a decrease in
accuracy. One may need to fine-tune the repair, match and fitness
functions for linguistic terms. However, we find the abstraction of
evo-devo mechanism is very apt to use.

\subsection{Machine Learning experiments}
In the machine learning technique of adaptation, 
\mycomment{we performed two
\myadd{sets of} experiments, namely,  
(a) the first experiment }
we considers $(\E, \K)$
pairs as the predominant entity and tests \myadd{the} accuracy of classification of
errors.
\mycomment{(b) the second experiment considers \myadd{the complete} sentence as the predominant
entity and measures \myadd{the} accuracy of corrections at the sentence level.}

In \mycomment{the first} \myadd{our} experiment, we used a total of $570$ misrecognition errors (for example,
\verbit{(dear, beer)} and \verbit{(have, has)} derived from $(\E_1, \K_1)$ or \verbit{(than twenty, jewelry)} 
derived from $(\E_2, \K_2)$)
in the $486$
sentences. We performed $10$-fold cross validation, 
each fold
containing 
$513$ 
$(\E, \K)$ 
pairs for 
training and
$57$ 
pairs for testing, 
Note that we assume the erroneous words in the ASR output being marked
by a human oracle, in the training as well as the testing set.
Suppose the following example ($\K_4$) occurs in the training set:
$$\K_4: \myboxnp{Which business has posted cumulative sales of
more than one million dollars from the 2007 to 2012}$$
$$\E_4: \myboxnp{which business has posted latest stills of
more than one million dollars from 2007 to 2012}$$
The classifier is given the 
pair 
$\{F$
(\verbit{latest stills}),
\verbit{cumulative sales}\} to the classifier. 
And if the following example occurs in the testing set ($\K_5$),
$$\K_5: \myboxnp{Did sales remain the same in retail between
two thousand thirteen and two thousand fourteen}$$
$$\E_5: \myboxnp{sales wine same in retail between two thousand
thirteen and two thousand fourteen}$$
the trained model or the classifier is
provided $F$(\verbit{wine}) and
successful repair would mean it correctly labels (adapts) it to
\verbit{remain the}.
The features used for classification were ($n=6$ in Equation (\ref{eq:feature_expand}))

\begin{center}
\begin{tabular}{ccl}
$f_{\beta 1}$ &$\rightarrow$&Left context (word to the left of $\E$), \\
$f_{\beta 2}$ &$\rightarrow$&Number of errors in the entire ASR sentence, \\
$f_{\beta 3}$ &$\rightarrow$&Number of words in $\E$, \\
$f_{\beta 4}$ &$\rightarrow$&Right context (word to the right of $\E$), \\
$f_{\beta 5}$ &$\rightarrow$&Bag of vowels of $\E$ and \\
$f_{\beta 6}$ &$\rightarrow$&Bag of consonants of $\E$.
\end{tabular}
\end{center}

The combination of features 
$f_{\beta 6}$, $f_{\beta 5}$, $f_{\beta 1}$, $f_{\beta 3}$ , $f_{\beta 4}$ namely, 
(bag of consonants, bag of vowels, left context, number of words, right context) 
gave the best results with
$32.28$\% \myadd{improvement in} accuracy in classification over $10$-fold validation.

\mycomment{
In the second experiment, we used $214$ sentences which had only
misrecognition errors, a total of $287$ errors. We used our classifier on
these sentences by performing a leave-one-out cross validation, since
we had limited data. These experiments were performed for every
combination of features as before, and the best results were found with
the combination (bag of consonants, bag of vowels, left context, right
context). Figure \ref{fig:1} shows the histogram plot of the improvement in
percentage accuracy in repaired ASR output based on the ML technique
for this feature combination. Of the $214$ runs, only one instance showed
a degradation of the ASR output (See Figure \ref{fig:1}). In the rest of the
cases, there was an improvement in accuracy of the repaired sentence
(based on word error rate) in $120$ runs, and in $93$ cases, there was no
change in sentence accuracy. The maximum absolute improvement for a
sentence was $17.6$\%.

\begin{figure}
\centering
\includegraphics[width=0.65\textwidth]{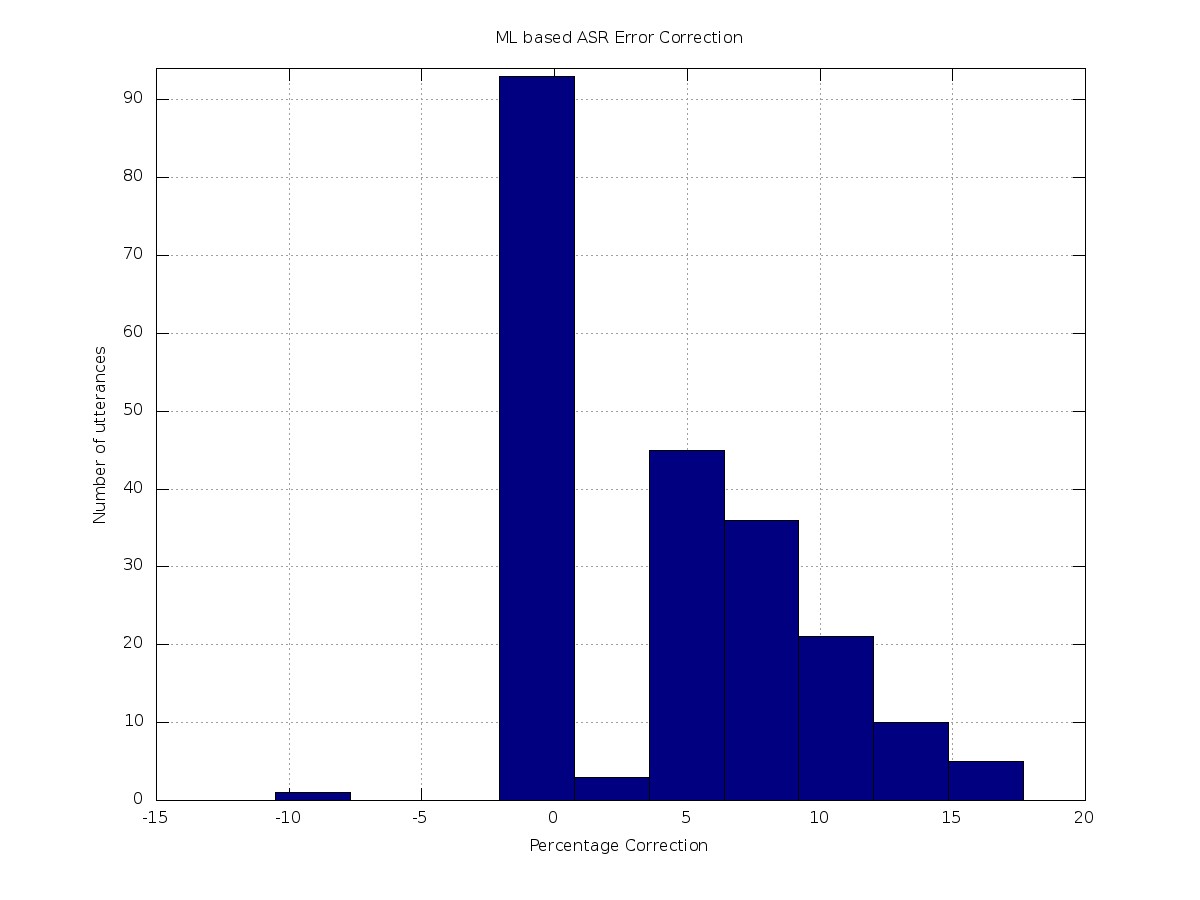}
\caption{Histogram showing percentage improvement by machine learning
adaptation}
\label{fig:1}
\end{figure}
}

\myadd{The experimental results for both evo-devo and machine learning based 
approaches demonstrate that these techniques can be used to correct the erroneous output of ASR. This is what we set out to establish in this paper.}

\mycomment{indicate that even with a small dataset, there are indeed
improvements to be gained. Future work will also include bigger
datasets that will hopefully lead to better machine learning.}

\section{Conclusions}
\label{sec:conclusions}

General-purpose ASR engines when used for enterprise domains may output
erroneous text, especially when encountering domain-specific terms. One
may have to adapt/repair the ASR output in order to do further natural
language processing such as question-answering. We have presented two
mechanisms for adaptation/repair of ASR-output with respect to a
domain. The Evo-Devo mechanism provides a bio-inspired abstraction to
help structure the adaptation and repair process. This is \myadd{one of} 
the main
contribution of this paper. The machine learning mechanism provides a
means of adaptation and repair by examining the feature-space of the
ASR output. The results of the experiments show that both these
mechanisms are promising and may need further development.

\section{Acknowledgments}
Nikhil, Chirag, Aditya have contributed in conducting some of the experiments.
We acknowledge their contribution.

\bibliographystyle{unsrtnat}
\bibliography{lrev_anantaram}

\end{document}